%% file: main.tex
\documentclass{article}
\usepackage{utils/spconf,amsmath,epsfig}

\usepackage{algorithmic}
\usepackage[ruled,vlined]{algorithm2e}
\usepackage[normalem]{ulem}
\usepackage{amssymb}
\usepackage{graphicx}
\usepackage{tabularx}
\usepackage{multirow}
\usepackage{booktabs}
\usepackage{subcaption}
\usepackage{enumitem}

\usepackage{float}
\usepackage{hyperref}
\hypersetup{
    colorlinks=true,
    linkcolor=red,
    filecolor=magenta,      
    urlcolor=cyan,
    }
\usepackage{comment}
\usepackage{tikz}
\let\OLDthebibliography\thebibliography
\renewcommand\thebibliography[1]{
  \OLDthebibliography{#1}
  \setlength{\parskip}{0pt}
  \setlength{\itemsep}{0pt plus 0.3ex}
}

\begin{document}\sloppy

% Example definitions.
% --------------------
\def\x{{\mathbf x}}
\def\L{{\cal L}}

% \newcommand\mycommfont[1]{\footnotesize\ttfamily\textcolor{blue}{#1}}
% \SetCommentSty{mycommfont}

\newcommand{\etal}{\textit{et al.}}
\newcommand{\steph}[1]{\textcolor{blue}{Steph: #1}}
\newcommand{\goluck}[1]{\textcolor{red}{Goluck: #1}}
\newcommand{\gv}[1]{\textcolor{magenta}{GV: #1}}
\definecolor{kpd}{HTML}{FF6666}
\definecolor{mtn}{HTML}{FF6666}

% Title.
% ------
\title{Predictive Coding for Animation-Based Video Compression}
%
% Single address.
% ---------------
\name{Goluck Konuko$^{\dagger}$, St\'ephane Lathuili\`ere$^{\ddagger}$, Giuseppe Valenzise$^{\dagger}$}
\address{$^{\dagger}$ Université Paris-Saclay, CentraleSupélec,  Laboratoire des signaux et systèmes\\
$^{\ddagger}$ LTCI, Télécom Paris, Institut Polytechnique de Paris, France}

\maketitle

\input{abstract.tex}
\input{introduction.tex}

\input{related.tex}

\input{method.tex}

\input{experiments.tex}

\input{conclusions.tex}

\newpage
\bibliographystyle{utils/IEEEbib}
\bibliography{refs}

\end{document}

%% file: abstract.tex
\begin{abstract}
% Animation-based video compression as applied to the video conferencing content includes compact motion representation using landmarks and the application of autoencoder networks to reconstruct target videos. Previous works have applied adversarial training to realize coding frameworks that deliver significantly higher perceptual quality in output videos at ultra-low bitrates relative to state-of-the-art video coding tools. However, residual coding which is a classical element of practical video compression has not been included into previous proposals for animation-based coding. Hence in this work, we propose a framework that jointly learns to animate faces from a single reference and a compact latent representation for the residual difference between the animated frame and the original frame. Further we constrain the animation model learn the temporal correlations on the residual layer to minimize the effective entropy of the residual layer bitstream. We present the quantitative evaluation metrics showing bitrate gains in excess of 70\% relative to HEVC and over 30\% relative to VVC for perceptual and pixel-based metrics.

We address the problem of efficiently compressing video for conferencing-type applications. We build on recent approaches based on image animation, which can achieve good reconstruction quality at very low bitrate by representing face motions with a compact set of sparse keypoints. However, these methods encode video in a frame-by-frame fashion, i.e., each frame is reconstructed from a reference frame, which limits the reconstruction quality when the bandwidth is larger. Instead, we propose a predictive coding scheme which uses image animation as a predictor, and codes the residual with respect to the actual target frame. The residuals can be in turn coded in a predictive manner, thus removing efficiently temporal dependencies. Our experiments indicate a significant bitrate gain, in excess of 70\% compared to the HEVC video standard and over 30\% compared to VVC, on a dataset of talking-head videos.
% We propose an animation-based video compression framework that uses compact motion representation using sparse motion keypoints and autoencoder networks for image compression and reconstruction. Previous works have used adversarial training of image animation models to achieve low bitrate compression of talking head videos with high perceptual quality. However these methods have not included residual coding, a classical element of practical video compression that allows for quality scalability. Our framework jointly learns to animate faces and encodes a compact residual difference information between the animated frame and the original frame. Our framework further learns the temporal correlations on the residual layer, which minimizes the effective entropy of the residual layer bitstream. We present quantitative evaluation metrics that demonstrate bitrate gains of over 70\% relative to HEVC and over 30\% relative to VVC for both perceptual and pixel-based metrics.

\end{abstract}

%% file: introduction.tex
\section{Introduction}
\label{sec:intro}
Recent work on learning-based video coding for videoconferencing applications has shown that it is possible to compress videos of talking heads with extremely low bitrate, without significant losses in visual quality~\cite{konuko2020dac,konuko2022hdac,wand2021conferencing,oquab2020low, chen2022beyondkp,chen2022som}. The basic tenet of these methods is that face motion can be represented through a compact set of sparse keypoints~\cite{siarohin2019first}, which can be transmitted and used at the decoder side to animate a reference video frame. 

However, despite the impressive coding performance of these methods at very low bitrates, existing animation-based codecs for videoconferencing still have several bottlenecks. Firstly, when the available bitrate increases, the reconstruction quality quickly reaches saturation, and conventional coding tools such as HEVC or VVC perform better. Secondly, bitrate variability in current schemes is complex, unlike conventional coding methods where a simple quantization parameter can be used to regulate bitrate. Finally, animation-based codecs operate on a frame-by-frame basis, which is inefficient for eliminating temporal redundancy in the video.

\begin{figure*}[t]
	\begin{center}
 \includegraphics[width=0.8\linewidth]{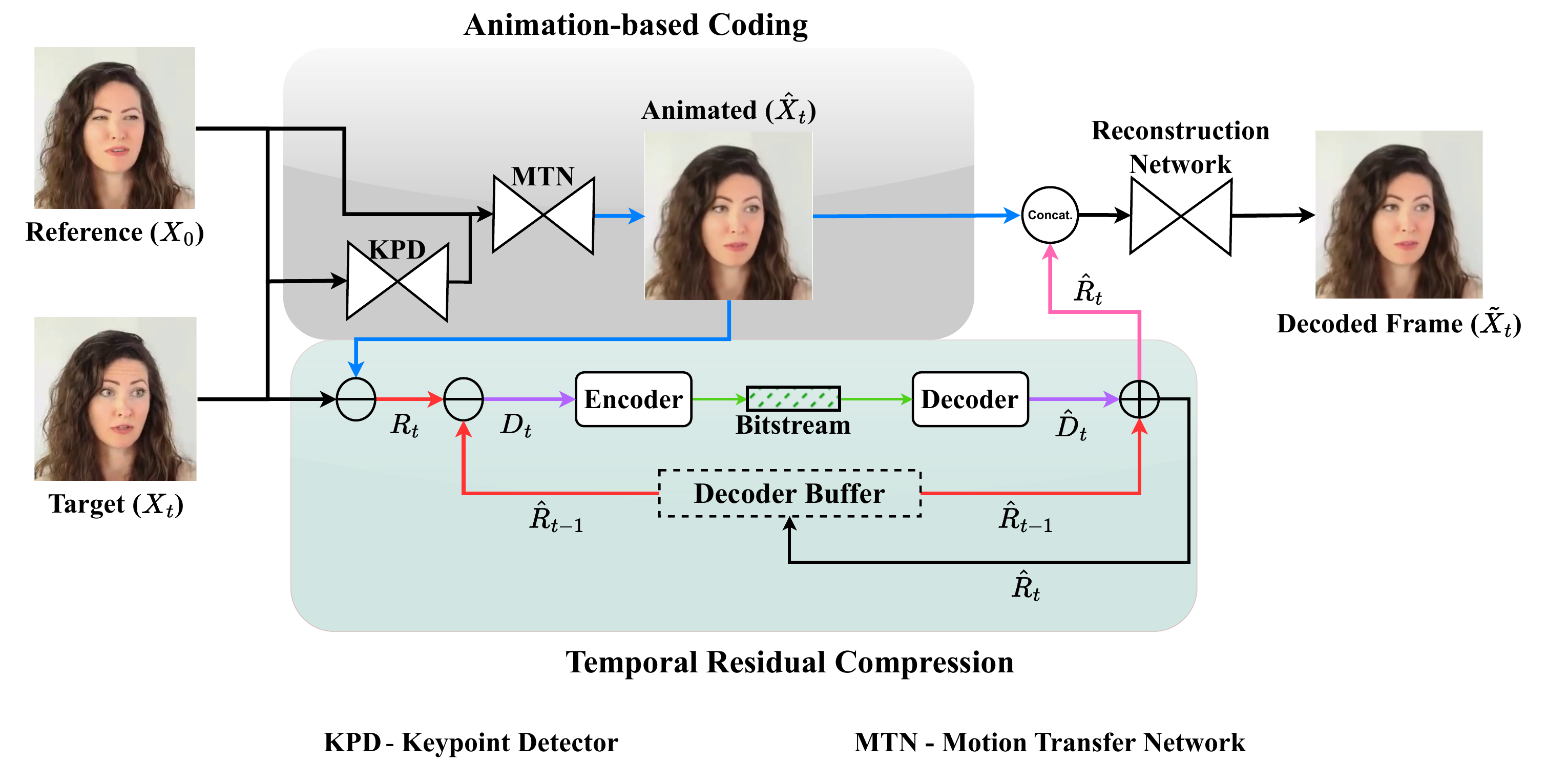}
	\end{center}
    \vspace{-0.25cm}	\caption{\textbf{Proposed RDAC framework}. From a single reference frame, motion keypoints and a compact residual layer, our framework reconstructs a video with high perceptual and pixel fidelity.
    Note that when previously decoded frame information is available, it is used to improve the prediction accuracy of the next target residual through temporal correlation learning.}
	\label{fig:pipeline}
\end{figure*}

This paper addresses these limitations by proposing a \textit{predictive coding} scheme for videoconferencing applications. Specifically, we interpret the keypoint-based image animation used in previous codecs~\cite{konuko2020dac} as a \textit{spatial predictor} of the current (target) frame, as depicted in Figure~\ref{fig:pipeline}. The residual between the animated and the target frame is then coded and used at the decoder side to correct the animated target frame. Since animation residuals exhibit \textit{temporal} correlation, we also encode them in a predictive manner, i.e., we predict the current animation residual based on the previously decoded residual and encode the prediction difference. 
It is worth noting that this approach is similar in principle to the classic video coding prediction loop, with the important distinction that residual coding and animation are \textit{jointly} learned in an end-to-end fashion.
We name our method \textbf{RDAC}, for Residual Deep Animation Codec.
Our results demonstrate significant rate-distortion improvements compared to standard codecs such as HEVC and VVC, as measured by several classical and learning-based perceptual quality metrics. Furthermore, the proposed technique has the additional advantage of reducing temporal drift compared to previous frame-by-frame approaches.

%% file: related.tex
\section{Related Work}
\label{sec:related}
Image animation models have been applied to compress talking head videos at ultra-low bitrates in conferencing-type applications~\cite{konuko2020dac,konuko2022hdac,wand2021conferencing,oquab2020low, chen2022beyondkp,chen2022som}. Different from other learning-based compression frameworks~\cite{rippel2017real,balle2018variational,agustsson2019generative,hu2021fvc,balle2016end,lu2019dvc,mentzer2022vct}, the animation-based codecs in \cite{wand2021conferencing} and~\cite{oquab2020low} propose architectures that use a variable number of motion keypoints to change the reconstruction quality within a small range of low bitrates. The deep animation codec (DAC) in our previous work~\cite{konuko2020dac} offers the possibility to vary the bitrate by creating a list of reference frames from which the best reconstruction is computed. Specifically, a new reference frame is added to the decoder buffer if all the available frames give reconstruction below a predefined threshold. However, this approach may introduce temporal jittering when adjacent animated frames are predicted from different reference frames. Using second-order motion coherence~\cite{chen2022som} introduces spatio-temporal stability in the decoded video, hence reducing the jittering. However, this architecture is still limited in terms of quality variability since it relies only on face animation. In our recent work~\cite{konuko2022hdac}, we proposed a hybrid coding architecture (HDAC) that uses a low-quality HEVC bitstream as side information to enhance the final result of the animation codec. While improving on previous methods, the use of this low-quality auxiliary stream limits in practice the possibility to reconstruct high-frequency details.

In this work, we propose a residual deep animation codec (RDAC) that learns a compact representation of the residual between a frame and its animation-based prediction, and encodes this residual using temporal prediction.

%% file: method.tex
\section{PROPOSED METHOD}
\label{sec:method}
A general scheme of the proposed residual deep animation codec is depicted in Fig.~\ref{fig:pipeline}. The components of the proposed system are detailed as follows: Section~\ref{subsec:residual-coding} introduces the frame prediction and residual coding and Section~\ref{subsec:temporal-residual} presents temporal learning in the residual space.
% At the encoder side, the animation-based prediction estimates a set of sparse keypoints to describe motion between a reference and a target frame. These keypoints are transmitted as part of the bitstream, and are used at the decoder side to animate the reference frame into a prediction of the target frame.
% The residual between the animation-based prediction and the target frame is computed and coded through the temporal residual compression module. 

\subsection{Deep Image Animation Prediction and Residual Coding}
\label{subsec:residual-coding}
We leverage the principles developed in the First Order Model~\cite{siarohin2019first} for image animation and our prior works~\cite{konuko2020dac, konuko2022hdac} for animation-based prediction. The image animation process works by estimating a sparse set of motion landmarks using a keypoint detector (KPD) which is a UNet-like architecture from~\cite{siarohin2019first}. The keypoints are used by a motion transfer network (MTN) that generates the optical flow between a decoded reference image $\mathbf{\tilde{X}}_{0}$ and the desired target $\mathbf{X}_{t}$. Subsequently, the optical-flow map is applied to the feature space representation of the reference frame  derived by the encoder of an autoencoder network. The deformed source features are assumed to be a close approximation of the target frame's feature representation and are used by a decoder network to produce the final animation $\mathbf{\hat{X}}_{t}$.

We build on this animation framework by including an encoder network that learns a latent representation of $\mathbf{R}_{t} = \mathbf{X}_{t} - \mathbf{\hat{X}}_{t}$ \textit{i.e.} the residual after animation as illustrated in Fig.~\ref{fig:pipeline}. We start with the architecture of the variational autoencoder network ~\cite{balle2018variational} used for learned image compression frameworks. However, since the residual images have very sparse features we mitigate the potential encoding of a noisy latent representation by increasing the number of downsampling convolutional layers from 3 to 5 and symmetrically increase the number of upsampling layers.

\subsection{Using Temporal Correlation in the Residual Layer}
\label{subsec:temporal-residual}
For a sequence of target frames $\mathbf{X}_{1} \to \mathbf{X}_{T}$ animated from a single reference frame, $\mathbf{X}_{0}$, we observe that the residual differences $\mathbf{R}_{1} \to \mathbf{R}_{T}$ have a high temporal correlation. In this paper, we use a simple differential coding scheme to exploit this temporal correlation. Specifically, we compute the temporal difference signal between consecutive frame residuals, $\mathbf{D}_{t} =  \mathbf{R}_{t}-\mathbf{\hat{R}}_{t-1}$, as shown in Fig.~\ref{fig:pipeline}. Note that, in general, more sophisticated prediction schemes are possible, that could bring additional temporal decorrelation, e.g., any dense or block-based motion compensated scheme. In this work, we demonstrated coding gains even with a suboptimal zero-motion temporal predictor, leaving the study of more advanced prediction schemes to future work. 

The difference signal $\mathbf{D}_{t}$ is coded using an additional autoencoder network, which is trained together with the animation-based predictor and the reconstruction network. The decoding process consists in reconstructing the residual $\mathbf{\tilde{R}}_t=\mathbf{\tilde{D}}_t + \mathbf{\tilde{R}}_{t-1}$. The reconstructed residual is then concatenated to the animation-based predictor $\mathbf{\hat{X}}_t$ and passed as input to a reconstruction network that produces the final decoded frame $\mathbf{\tilde{X}}_t$. The reconstruction network consists of 2 convolution layers and 3 ResNet blocks.

\subsection{Model Training}
\label{subsec:training}
 We initialize the animation module with pre-trained models from~\cite{konuko2020dac}. The loss terms for image animation are the same as in~\cite{konuko2020dac, siarohin2019first}, while the rate-distortion loss $\mathcal{L}_{RD}$ is derived as described in~\cite{balle2018variational}:
\begin{equation}
   \mathcal{L}_{RD} = \lambda \cdot \text{MSE}(\mathbf{R_t}, \mathbf{\hat{R}_t}) +  \text{Rate}
\end{equation}
where the bitrate cost in bits-per-pixel (bpp) is computed from the entropy estimate of the residual latent representation. 

\begin{figure*}[t]
 \def\im#1{ \includegraphics[width=27mm,height=27mm]{#1}}
 %\vspace{0.3cm}
     \centering
   \setlength\tabcolsep{0.5 pt}
   \renewcommand{\arraystretch}{0.2}
     \begin{tabular}{*{7}{c}}
    \toprule
    Ground Truth  & HEVC (8 kbps) & VVC (8kbps)  & DAC (4.4 kbps) & HDAC (10kbps) & \textcolor{red}{RDAC (Ours) (8 kbps)} \\
    \midrule
    \im{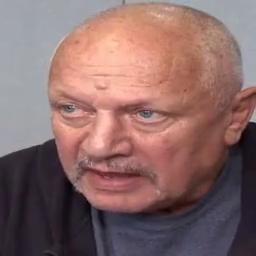} &
    \im{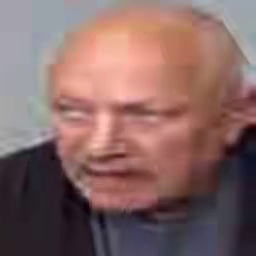} &
    \im{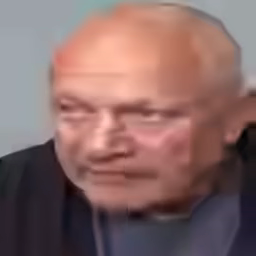} &
    \im{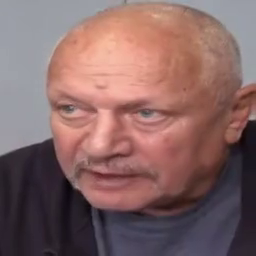} &
    \im{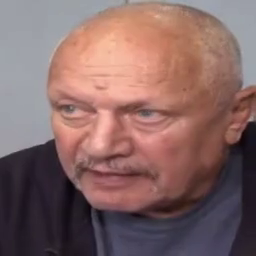} &
    \im{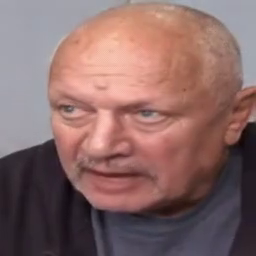} 
    
    \\
    
    \im{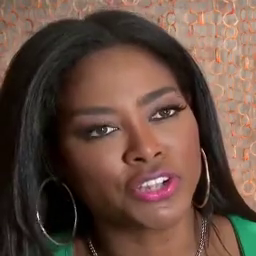} &
    \im{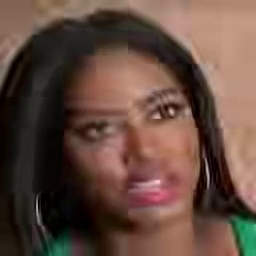} &
    \im{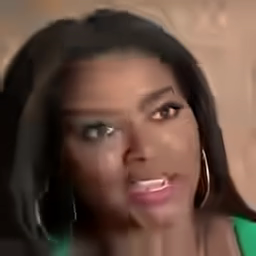} &
    \im{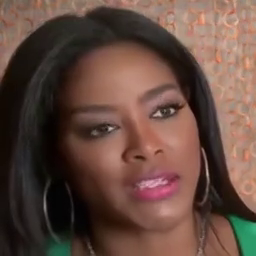}&
    \im{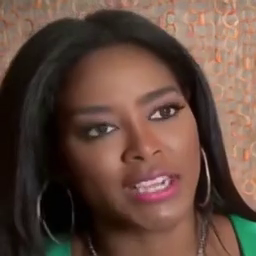}&
    \im{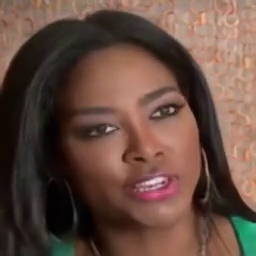}
   
     \end{tabular}
\caption{\textbf{Visual Comparison of coding results.} A qualitative comparison of our proposed coding framework shows significant quality improvement over HEVC and VVC at low bitrates. We observe fewer smoothing and blocking artifacts as well as better color and style preservation in the reconstructed frames with our framework.}
\label{fig:visual-comparison}
\end{figure*}

%% file: experiments.tex
\section{Experiments and Results}
\subsection{Evaluation Protocol}
\label{subsec:training}
  We randomly select 30 video sequences from the VoxCeleb test set with minimum lengths of 128 frames. We note that chaning the GOP size affects the average reconstruction quality of the video sequences. Therefore, we encode the sequences with GOP sizes 16, 32, 64, and 128 and select the best reconstruction point at each bitrate from a union of the computed metrics \textit{i.e.} the convex hull of all the GOP configurations. The reference frame is encoded with QP 30 using the BPG codec (HEVC intra) and the motion keypoints as well as the compressed residuals are entropy coded using a context-adaptive arithmetic coder with a Prediction by Partial Match (PPM) model~\cite{cleary1984ppm}. HEVC and VVC (VTM-11) metrics are computed under low-delay configurations with high QP values to minimize bitrate. We also compare against the LPIPS-VGG metrics reported for BeyondKP~\cite{chen2022beyondkp} and FaceVid2Vid~\cite{wand2021conferencing} since they use comparable test conditions. Notice that for these last two methods, we only have a single bitrate point, since they do not support bitrate variability beyond 10 kbps. MSE loss is used at training time for residual learning. However, the other loss terms used in training the network optimize for perceptual quality. Therefore, we restrict our evaluation to use only perceptual metrics and multi-scale pixel fidelity metrics.
  
\subsection{RD Evaluation}
\label{subsec:rd-evaluation}
\begin{figure}[h]
    \centering
    \includegraphics[width=7cm]{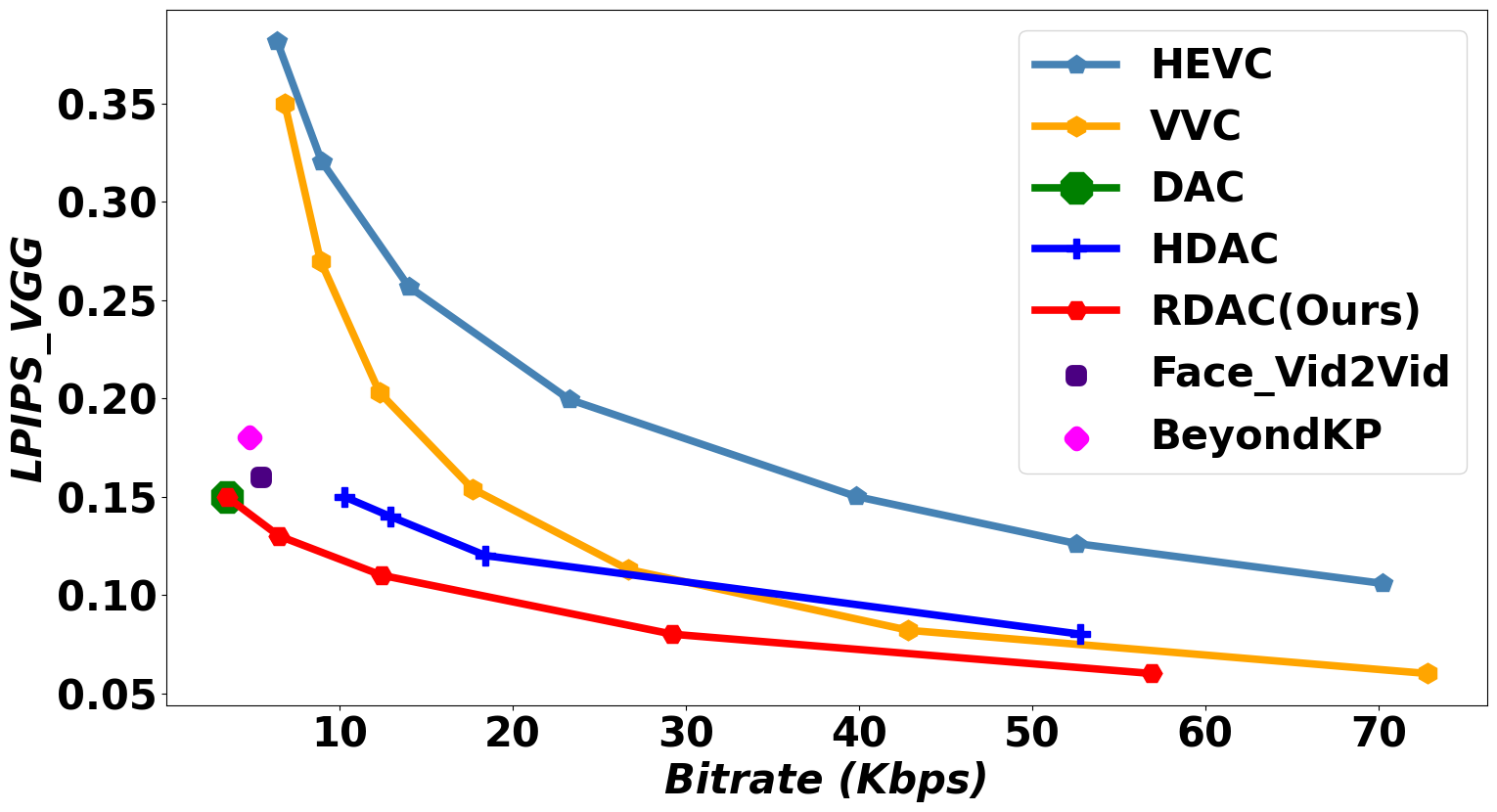}
    \caption{\textbf{RD Performance:} Our method achieves considerable bitrate gains over a wider range of bitrates relative to both state-of-the-art video compression tools and the previously proposed frameworks.}
    \label{fig:rd-sota}
\end{figure}

In Tab.~\ref{tab:bd-gains}, we note over 70\% bitrate savings for perceptual-based metrics \textit{i.e.} LPIPS~\cite{zhang2018lpips}, msVGG~\cite{simonyan2015vgg} and DISTS~\cite{ding2021dists} as well as over 40\% bitrate savings for pixel-based metrics over HEVC. In Fig.~\ref{fig:visual-comparison} we make a visual comparison of our proposed framework with HEVC and VVC in the low bitrate range.
\begin{table}[H]
\small
\begin{center}
\caption{\textbf{Bitrate savings (\% BD-BR)} computed over 30 video sequences with 128 frames from VoxCeleb test set. The bitrate savings are measured with HDAC~\cite{konuko2022hdac}, HEVC and VVC codec as anchors}

\begin{tabular}{*{4}{c}}
\toprule
% &\multicolumn{3}{c}{\textbf{BD Bitrate Savings}}\\
\textit{\textbf{Metrics}} &{\textit{\textbf{HDAC}}}&\textit{\textbf{VVC}}& \textit{\textbf{HEVC}}  \\ 
\midrule
    \textbf{msVGG} & -55.10  & -66.65 & -74.99  \\
    \textbf{DISTS}  & -48.46 & -63.01  & -82.62  \\
    \textbf{LPIPS-VGG}  & -38.43 & -48.73 & -78.96  \\
    \textbf{LPIPS} & -6.45  & -33.11 & -75.96 \\
    \textbf{FSIM}  & -42.89 & -16.02 &  -63.10 \\
    \textbf{MS-SSIM}  & -56.97   & -20.11  &  -52.05 \\
    % \textbf{IW-SSIM}  & -51.51 & -37.22 & -40.02  \\
  \bottomrule
\end{tabular}
\label{tab:bd-gains}
\end{center}
\vspace{-0.5cm}
\end{table}
Fig.~\ref{fig:rd-sota} illustrates the rate-distortion performance using the LPIPS metric. RDAC significantly improves performance of conventional video codecs over a wide range of bitrates, and it outperforms previous animation-based codecs which do not employ predictive coding. 

\subsection{Ablation study and temporal drift}
\label{subsec:ablation}
\begin{figure}[t]
    \centering
    \includegraphics[width=7cm]{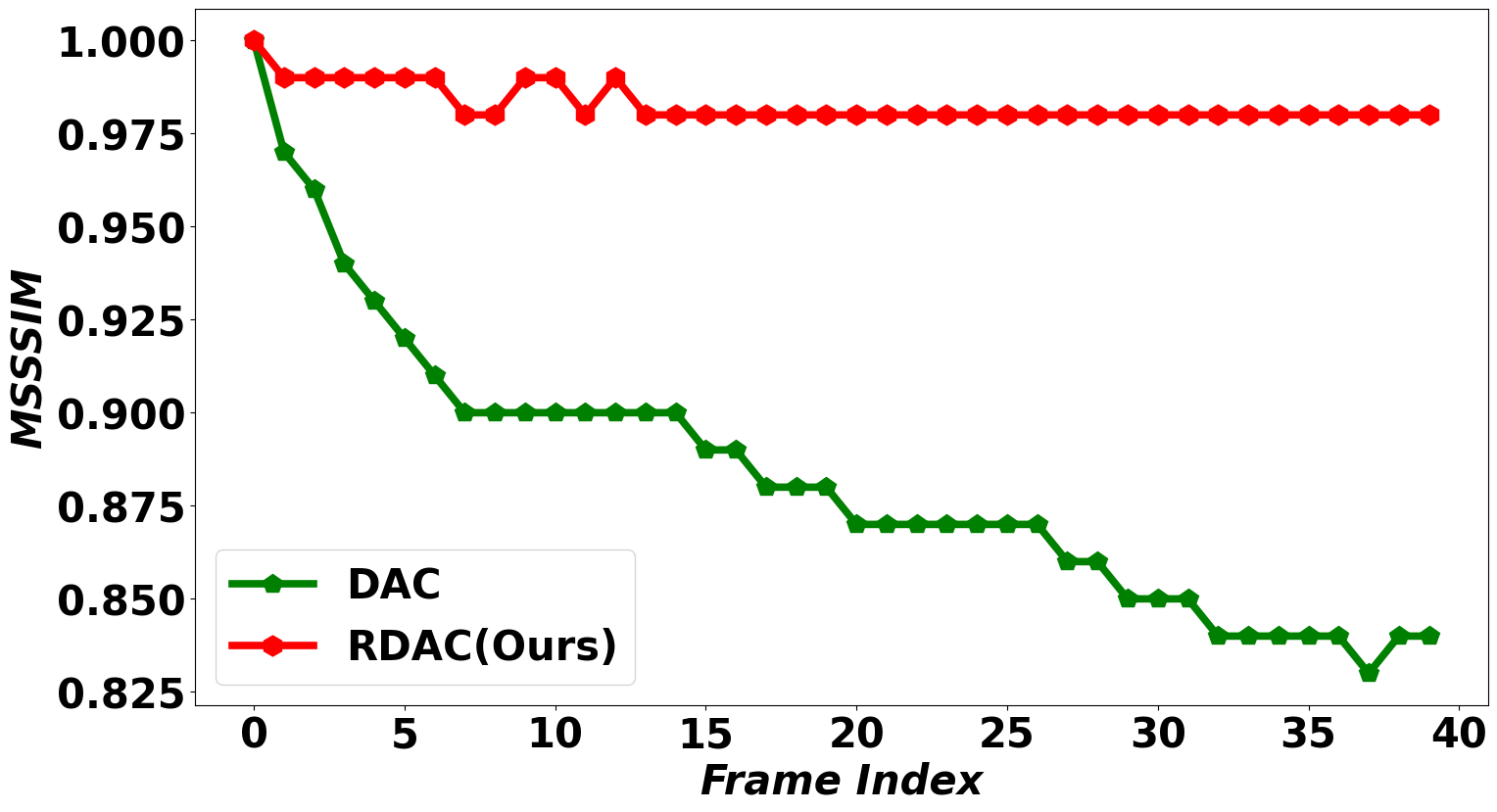}
    \caption{\textbf{Reconstruction quality as a function of time:} RDAC temporal prediction avoids temporal drift, in contrast with open-loop schemes such as DAC, where quality degrades as the target frames get farther from the reference.}
    \label{fig:temporal-drift}
\end{figure}

An advantage of using a closed-loop prediction scheme for temporal coding of residuals is that it avoids the temporal drifting affecting previous open-loop schemes such as DAC. This is supported by Fig.~\ref{fig:temporal-drift}, where we show the temporal reconstruction quality (measured with MS-SSIM) of our framework and DAC. 

We also investigate to which extent the temporal prediction contributes to the RD gains, over a frame-by-frame scheme to code the prediction residuals $\mathbf{R}_t$. To this end, we remove the temporal feedback loop in Fig.~\ref{fig:pipeline}, encoding the residuals as all Intra. Tab.~\ref{tab:bd-gains-rdac} reports the gains of our proposed RDAC (with temporal prediction) over this simpler solution, demonstrating that reducing temporal correlation has a significant impact on coding performance.
\begin{table}[ht]
% \small
\begin{center}
\caption{\textbf{Bitrate Savings (\% BD-BR)} from our framework with temporal residual learning versus no temporal residual (10 sequences/64 frames)}

\begin{tabular}{*{5}{c}}
\toprule
\textit{DISTS}&\textit{LPIPS} & \textit{msVGG} &\textit{FSIM} & \textit{MS-SSIM} \\ 
\midrule
-9.35 & -14.32 & -5.65 & -9.20 & -6.86\\
\bottomrule
\end{tabular}
\label{tab:bd-gains-rdac}
\end{center}
\vspace{-0.5cm}
\end{table}

\subsection{Computational complexity}
\label{subsec:complexity}
In Tab.~\ref{tab:complexity}, we make a complexity evaluation by comparing the coding or decoding time for a single interframe. The animation-based models DAC, HDAC, and our framework are evaluated on a CPU and GPU while the HEVC and VVC codecs are only evaluated on a CPU since they do not have GPU acceleration capability. We note that our proposal adds only a moderate level of complexity relative to HEVC. However since we achieve bitrate savings greater than VVC, we consider this additional complexity as an acceptable tradeoff for the target application.
\begin{table}[H]
\small
\begin{center}
\caption{\textbf{Computational Complexity:} Time to encode/decode 1 frame (in seconds). The computation time is estimated for the highest RD point of our framework. }

\begin{tabular}{*{5}{c}}
\toprule
&\multicolumn{2}{c}{\textbf{CPU (Intel Corei7) }}& \multicolumn{2}{c}{\textbf{GPU (RTX 3090)}} \\
&{\textit{Enc.}}&\textit{Dec.}&{\textit{Enc.}}&\textit{Dec.}\\ 
\midrule
    \textbf{HEVC} & 0.09 & 0.005 & - & -\\
    \textbf{VVC} & 13.5 & 0.01 & -& -\\
\midrule
    \textbf{DAC} & 0.04 & 0.35 & 0.03 & 0.02 \\
  
    \textbf{HDAC} & 0.10 & 0.35 & 0.12 & 0.02 \\

    \textbf{\textcolor{red}{RDAC (Ours)}}  & 0.52 & 0.50 & 0.21 & 0.02 \\
  \bottomrule
\end{tabular}
\label{tab:complexity}
\end{center}
\vspace{-0.5cm}
\end{table}

%% file: conclusions.tex
\section{Conclusions}
Animation-based compression offers the possibility to transmit videos with very low bitrate. However, it is often limited to reconstructing the outputs at a fixed quality level, cannot scale efficiently when higher bandwidth is available, and does not compress efficiently temporal redundancies in the signal. In this paper, we propose a coding scheme that integrates image animation (re-interpreted as a frame predictor) with classical predictive coding principles, where we exploit both spatial and temporal dependencies to achieve a coding gain. Our RDAC codec outperforms previous methods and standard codecs by a large margin on a dataset of talking head videos, despite the very simple temporal prediction approach employed.
% Future work will address more efficient temporal predictors to further improve coding performance.

\textbf{Acknowledgement:} This work was funded by Labex DigiCosme - Universit\'e Paris-Saclay. This work was performed using HPC resources from GENCI-IDRIS %(Grant 2021-AD011013152)